\documentclass[letterpaper]{article} 
\usepackage[preprint]{aaai2027} 
\usepackage[hyphens]{url}  
\usepackage{graphicx} 
\usepackage{natbib}  
\usepackage{caption} 
\usepackage{algorithm}
\usepackage{url}            
\usepackage{booktabs}       
\usepackage{amsfonts}       
\usepackage{nicefrac}       
\usepackage{microtype}      
\usepackage{xcolor}  
\usepackage{float}
\usepackage{algpseudocode}
\usepackage{amsmath} 
\usepackage{graphicx}
\usepackage{multirow}
\usepackage{colortbl}
\usepackage{xcolor}
\definecolor{posgreen}{RGB}{198,239,206}
\definecolor{negred}{RGB}{255,199,206}
\newcommand{\posc}[1]{\cellcolor{posgreen}+#1\%}
\newcommand{\negc}[1]{\cellcolor{negred}#1\%}

\usepackage{newfloat}
\usepackage{listings}
\DeclareCaptionStyle{ruled}{labelfont=normalfont,labelsep=colon,strut=off} 
\floatstyle{ruled}
\newfloat{listing}{tb}{lst}{}
\floatname{listing}{Listing}

\usepackage{booktabs}

\title{Catching the Imposter: Self-Supervised Learning of Physical Coherence with Cross-Entity Feature Permutations}
\author{
    Written by AAAI Press Staff\textsuperscript{\rm 1}\thanks{With help from the AAAI Publications Committee.}\\
    AAAI Style Contributions by Peter Patel Schneider,
    Sunil Issar,\\
    J. Scott Penberthy,
    George Ferguson,
    Hans Guesgen,
    Francisco Cruz\equalcontrib\corresponding,
    Marc Pujol-Gonzalez\equalcontrib\corresponding
}
\affiliations{
    \textsuperscript{\rm 1}Association for the Advancement of Artificial Intelligence\\

    1101 Pennsylvania Ave, NW Suite 300\\
    Washington, DC 20004 USA\\
    proceedings-questions@aaai.org
}

\title{Catching the Imposter: Self-Supervised Learning of Physical Coherence with Cross-Entity Feature Permutations}
\author {
    Aleksei Rozanov,
    Arvind Renganathan,
    Vipin Kumar
}
\affiliations {
    Department of Computer Science \& Engineering, University of Minnesota -- Twin Cities\\
    rozan012@umn.edu, renga@umn.edu, kumar001@umn.edu
}

\begin{document}

\maketitle

\begin{abstract}
Scientific data often describe entities whose features are jointly governed by the laws of physics, yet existing self-supervised learning (SSL) objectives largely ignore this physical coherence. We introduce \textit{imposter}, a discriminative pretext task that replaces subsets of an entity’s features with real observations donated by another entity and trains the encoder to identify the swapped features. Because every donated value is individually plausible, the task can only be solved by learning cross-feature physical dependencies. We evaluate the proposed objectives on global ERA5-Land reanalysis data using 21 environmental variables and assess the learned representations on seven downstream tasks spanning climate classification, carbon flux estimation, and streamflow prediction. Our study includes, to our knowledge, the first systematic comparison of self-supervised objectives for land-surface modeling under a shared architecture and pre-training budget. We find that the most effective pretext task depends on the downstream task family rather than any single objective’s superiority, and that \textit{imposter} provides complementary information when combined with existing SSL objectives. These results suggest that physical coherence is a valuable new source of self-supervision for scientific foundation models.
\end{abstract}


\section{Introduction}
Many scientific problems involve modeling entities with physically coupled features evolving in time: an ecosystem characterized by temperature, radiation, and water; a patient described by heart rate, blood pressure, and oxygen saturation; an engine instrumented with sensors measuring thermodynamically related parameters. This structure is especially pronounced in the Earth and environmental sciences, where processes central to climate monitoring, such as the terrestrial carbon and water cycles, are governed by tightly coupled physical variables processes \citep{xiao2026insights}. In these domains, labels are scarce and expensive, for e.g.  carbon flux measurements come from a few hundred eddy-covariance towers scattered across the globe \citep{pastorello2020fluxnet2015}, and streamflow records are available only where rivers are gauged, covering a small fraction of the world's catchments \citep{addor2017camels}.  Hence, the vast majority of the land surface remains unobserved. Meanwhile, unlabeled observations are abundant, with reanalysis products \citep{munoz2021era5} and satellite remote sensing providing decades of continuous global coverage. Self-supervised learning (SSL) is therefore a natural framing: it exploits the underlying structure of the unlabeled data to build expressive representations that improve downstream performance, with direct bearing on how well we can monitor carbon fluxes, water availability, and climate at a global scale.

\begin{figure}
    \centering
    \includegraphics[width=0.5\linewidth]{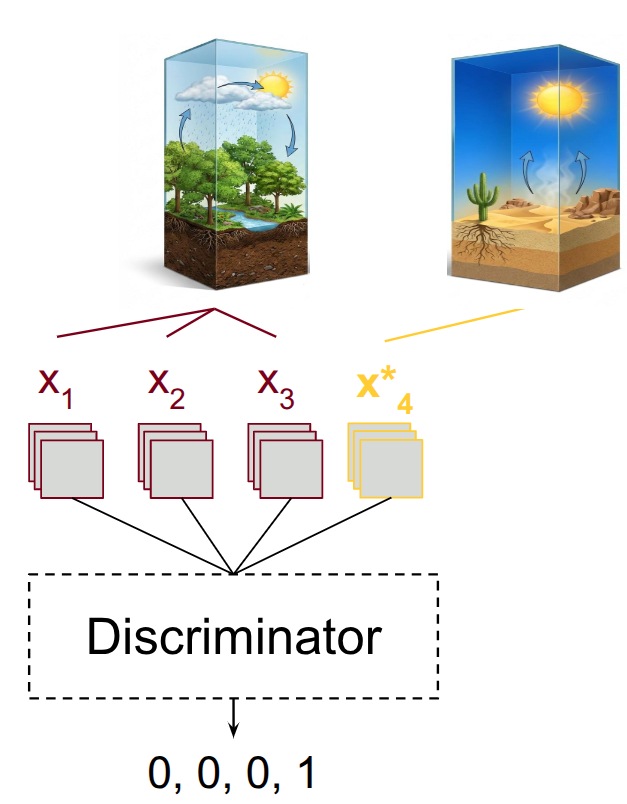}
    \caption{\small Overview of the \textit{imposter} pretext task. Given two entities representing temperate forest and desert ecosystems, the algorithm performs feature-level swap to detect incoherent features during the training stage.}
    \label{fig:main}
\end{figure}

SSL has driven major progress in natural language processing through next token prediction (NTP)~\citep{radford2019language, brown2020language} and bi-directional masked language modeling~\citep{devlin2019bert}, and in computer vision through masked autoencoding (MAE)~\citep{he2022masked, yuan2021florence} and contrastive learning~\citep{chen2020simple, tian2020contrastive}. Inspired by their success in language and vision, these ideas have naturally gained wide adoption in scientific domains dealing with colossal amounts of simulation data and real-world observations. Environmental modeling in particular has experienced rapid growth of self-supervised pre-training with objectives leveraging both the spatial and temporal structure of the data: masking has become ubiquitous in remote sensing foundation models~\citep{jakubik2023foundation, szwarcman2025prithvi}, while data-driven weather modeling~\citep{nguyen2023climax, lam2023learning, bodnar2025foundation, pathak2022fourcastnet} has converged on forecasting, the natural analog of NTP.

More broadly, self-supervised objectives derive supervision by exploiting structural properties inherent in the data. While these methods leverage the spatial and temporal regularity of the data, they remain agnostic to a defining property of scientific observation: the features of an entity are jointly governed by the laws of physics. More generally, scientific systems exhibit forms of structure that are largely absent from language and vision, including physical coherence among variables governed by shared processes. Existing objectives therefore exploit only part of the structure present in scientific data. Image augmentations such as cropping, flipping, and color jitter have no physical analog for a multivariate sensor record. Masking and forecasting treat each feature as an independent prediction target, so the model is never required to judge whether a \emph{combination} of features is physically consistent. Marginal-sampling corruption produces values that are plausible in isolation but, by design, breaks the joint dependence between variables and temporal autocorrelation. Yet it is precisely this physical coherence, with temperature, radiation, etc. co-varying as views of a single underlying system, that carries much of the information in scientific data.


To instantiate physical coherence as a source of self-supervision, we propose \textit{imposter} detection (Fig.~\ref{fig:main}), a discriminative pretext task that uses physical coherence itself as the supervisory signal. A subset of an entity’s features is randomly replaced with the corresponding features from a different entity, or in other words donated, and the model is trained on a per-feature binary classification objective to identify the inputs that break a physically coherent view of the receiving entity (i.e., the encoder is trained as a discriminator). Solving this task is only possible if the encoder learns the cross-feature dependencies that characterize the underlying system: because the donated features are real observations rather than synthetic corruptions, every replacement is individually plausible, and only its combination with the host’s remaining features reveals the swap. Thus, the corruption simultaneously provides supervision and serves as an efficient, physically grounded data augmentation. We additionally introduce \textit{multi-imposter}, a variant in which each swapped feature is drawn from an independent donor rather than a single one; this closes a shortcut of the base task in which the encoder can memorize the donor instead of evaluating each feature’s coherence with the host.

The idea of corrupting inputs with plausible replacements and training a discriminator to detect them has been explored in other domains. In language, ELECTRA~\citep{clark2020electra} replaces tokens with samples from a learned generator, while in the tabular setting, VIME~\citep{yoon2020vime} and SCARF~\citep{bahri2021scarf} corrupt features by sampling from each feature’s empirical marginal distribution. Our formulation adapts this general paradigm to multivariate scientific time series by replacing features with real observations donated from other entities rather than synthetic samples. Because every donated value is individually plausible, detecting the swap requires reasoning about cross-feature dependencies that unfold over time to determine whether each donated feature preserves the physical coherence of the host entity.We expand on these distinctions in the Related Work.


If different self-supervised objectives exploit different structural properties of scientific data, they should learn complementary representations. Accordingly, we hypothesize that the signal targeted by \textit{imposter} is complementary to those extracted by existing pretext tasks. Masking, forecasting, and contrastive learning primarily leverage spatial and temporal structure, whereas \textit{imposter} exploits cross-feature physical coherence. Because these sources of structure are largely orthogonal, the objectives are complementary rather than mutually exclusive. We therefore study whether combining \textit{imposter} with five widely used self-supervised pretext tasks in climate science, using both concatenated embeddings and joint training, yields richer representations than any objective alone. This allows us to quantify the unique contribution of each objective and assess whether their combination captures a broader spectrum of the structural information present in scientific data.

We conduct this study on the global daily ERA5-Land reanalysis \citep{munoz2021era5}, pre-training encoders on 21 features describing terrestrial energy, water, and vegetation processes, and assessing the learned representations on seven downstream tasks: K\"{o}ppen climate classification \citep{essd-13-5087-2021} (5- and 30-class targets), carbon flux estimation on a dedicated benchmark \citep{rozanov2026carbonbench} (GPP, RECO, NEE), and streamflow and basin identity prediction using the CAMELS dataset \citep{addor2017camels}.

Our study reveals that the choice of pretext task is governed by the downstream task family rather than any single objective's superiority, and that \textit{imposter}'s feature-coherence signal is complementary to existing objectives. Concretely, our contributions are as follows:

\begin{enumerate}
    \item We identify physical coherence as a previously underexplored source of self-supervision for multivariate scientific time series and instantiate this principle through the proposed \textit{imposter} and \textit{multi-imposter} pretext tasks.
    \item We systematically study how \textit{imposter} interacts with five widely used SSL objectives, via concatenated embeddings and joint training, benchmarking our method against and alongside each of them.
   \item We evaluate all objectives under a shared architecture and pre-training budget across seven climate-relevant downstream tasks, providing, to our knowledge, the first systematic comparison of SSL objectives for land-surface modeling.
\end{enumerate}

\section{Related Works}
\paragraph{Masking and reconstruction.}
Masked autoencoding (MAE) has become a dominant SSL paradigm across many modalities and domains. In NLP, BERT~\cite{devlin2019bert} and its extensions~\cite{liu2019roberta, joshi2020spanbert} pre-train transformers~\cite{vaswani2017attention} by masking input tokens and predicting them from context. In vision, masked autoencoders~\cite{he2022masked, xie2022simmim}, building on earlier inpainting formulations~\cite{pathak2016context, iizuka2017globally}, have matched or exceeded supervised pre-training and showed even better generalizazation with scaling. These methods share the same idea of corrupting the input by removing information, and then training the model to reconstruct it.

\paragraph{Contrastive learning.}
Rather than reconstructing corrupted inputs, contrastive methods~\cite{wang2015unsupervised, tian2020contrastive, hassani2020contrastive} learn representations by discriminating between similar and dissimilar pairs. SimCLR~\cite{chen2020simple} demonstrated that augmented views of the same image, treated as positive pairs within a batch, yield strong representations for downstream tasks. Contrastive objectives have since been extended to time series~\cite{tonekaboni2021unsupervised, franceschi2019unsupervised, yue2022ts2vec}, though these approaches typically rely on temporal augmentations rather than feature-level corruption.

\paragraph{Other pretext tasks and the AI4Science gap.}
Beyond masking and contrastive objectives, a variety of pretext tasks have been explored, e.g. jigsaw puzzles~\cite{noroozi2016unsupervised}, future prediction~\cite{srivastava2015unsupervised}, denoising~\cite{pang2021recorrupted}, and adversarial generation~\cite{goodfellow2020generative}. While these methods have driven progress in vision and language, their adoption in scientific domains remains limited. Scientific data poses distinct challenges: features are governed by physical laws, standard augmentations are inapplicable since entities are characterized by coupled multivariate time series rather than images or token sequences. Recent works have applied SSL to protein structure modeling~\cite{rives2021biological, lin2023evolutionary}, material property prediction~\cite{magar2022crystal}, molecular representation learning~\cite{liu2023molxpt}, and earth system modeling~\cite{brown2025alphaearth, ravirathinam2026towards, pathak2022fourcastnet}, but these approaches predominantly adapt existing vision or language objectives to domain-specific architectures.

\paragraph{Corruption-based discrimination.}
A separate line of work replaces input elements with plausible alternatives and trains the model to detect the replacements, or in other words, formulates a discriminative rather than generative objective. ELECTRA~\cite{clark2020electra} substitutes natural language tokens with outputs from a small generator network, training the main encoder as a classifier. The authors showed that for the same model size, data, and compute, ELECTRA learns contextual representations that significantly outperform those produced by BERT. In the tabular domain, VIME~\cite{yoon2020vime} and SCARF~\cite{bahri2021scarf} both corrupt features by sampling from the empirical marginal distribution of each feature. They differ in objective: VIME jointly optimizes feature reconstruction and corruption mask recovery, while SCARF applies a contrastive loss between original and corrupted views. Both demonstrate strong results on tabular benchmarks, but neither considers temporal structure nor cross-entity feature donation.

\section{Methodology}

Consider an unlabeled dataset of temporal entities $\mathcal{D} = \{x_i\}_{i=1}^{N}$, where each $x_i \in \mathbb{R}^{T \times F}$ represents an entity observed over $T$ time steps across $F$ features. Our goal is to learn an encoder $f_\theta$ that produces representations useful for downstream tasks with limited supervision.

\begin{algorithm}[t]
\small
\caption{Imposter Pretext Task}
\label{alg:imposter}
\begin{algorithmic}[1]
\Require Unlabeled temporal dataset $\mathcal{D} = \{x_i\}_{i=1}^N$, $x_i \in \mathbb{R}^{T \times F}$; swap ratio $r$
\Ensure Pretrained encoder $f_\theta$
\For{each mini-batch $\mathcal{B} = \{x_1, \dots, x_B\}$}
    \State $\tilde{x} \leftarrow \textsc{Permute}(\mathcal{B})$ \Comment{random donor for each sample}
    \State $M \leftarrow \mathbf{0}^{B \times F}$ \Comment{initialize swap mask}
    \For{$i = 1$ to $B$}
        \State $n \sim \text{Uniform}(1, r \cdot F),\ n \in \mathbb{Z}$ 
        \State $\mathcal{S}_i \leftarrow$ random subset of $\{1, \dots, F\}$ with $|\mathcal{S}_i| = n$
        \State $x_i[:, \mathcal{S}_i] \leftarrow \tilde{x}_i[:, \mathcal{S}_i]$ \Comment{replace with donor features}
        \State $M[i, \mathcal{S}_i] \leftarrow 1$
    \EndFor
    \State $\hat{M} \leftarrow f_\theta(X)$ \Comment{predict which features were swapped}
    \State Update $\theta$ by $\nabla \, \text{BCE}(\hat{M}, M)$
\EndFor
\State \Return $f_\theta$
\end{algorithmic}
\end{algorithm}

The \textit{imposter} (Algorithm~\ref{alg:imposter}) task trains an encoder to detect cross-feature incoherence within an entity. Given a batch of entities, we construct corrupted samples by replacing a subset of each entity's features with the corresponding features from a randomly assigned \textit{donor} entity. The encoder is trained to identify which features were replaced through a per-feature binary classification objective. Because donated features are real observations from a different entity, every individual feature value remains statistically plausible. Only the combination with the host entity's remaining features is physically incoherent. Detecting this incoherence requires the encoder to learn cross-feature dependencies that arise from shared physical processes --- for instance, the joint distribution of temperature, radiation, and moisture at a given location.

The same idea can be viewed probabilistically: the discriminator implicitly learns to weight joint distributions of original and corrupted instances differently. For a coherent entity $(x_1, \dots, x_n)$ and donor-swapped one $(\tilde{x}_1, x_2, \dots, x_n)$, we expect $
    p(x_1, x_2, \dots, x_n) > p(\tilde{x}_1, x_2, \dots, x_n),
$
where $\tilde{x}_i$ is a feature drawn from another entity. The pretext task provides supervision for this inequality.

The primary hyperparameter is the swap ratio $r$, which controls how many features are replaced per entity. By default, we sample the swap count uniformly at random from $[1, r \cdot F]$ per sample --- a schedule we find outperforms a fixed rate (Appendix~\ref{app:swaps}). Random scheduling prevents the encoder from exploiting the constant corruption rate as a shortcut.

The default \textit{imposter} algorithm lets the encoder solve the task by identifying features that are sampled from strictly one donor which can lead to "memorization" of the donor when swap ratio is high. To eliminate this shortcut, we introduce \textit{multi-imposter}, in which each swapped feature is drawn from an independent donor. This decouples the corrupted features from each other and forces the encoder to evaluate each feature's coherence with the host in isolation which we claim to be a more challenging task.

The imposter task belongs to the family of corruption-based SSL methods. Its closest analog is masked autoencoding (MAE): where MAE corrupts the input by removing information (zero-masking) and trains the model to reconstruct it, imposter substitutes information with realistic alternatives and trains the model to detect the substitution. This shift from generative reconstruction to discriminative detection has a direct consequence for what the encoder is required to learn. MAE-style objectives encourage the encoder to preserve enough information to reconstruct the masked signal, whereas the \textit{imposter} objective instead encourages it to model the cross-feature dependencies that distinguish coherent from incoherent configurations.
 

\section{Experimental Settings}  \label{sec:exp}

\subsection{Architecture}
We use the same patch transformer architecture across all methods to isolate the effect of the pre-training objective. The input time series $x \in \mathbb{R}^{T \times F}$ is divided into non-overlapping patches along the temporal axis using a 1D convolution with kernel size and stride equal to the patch length. Each patch is projected to a $d$-dimensional embedding and added to learnable positional encodings, producing a sequence of $T/P$ tokens, where $P$ is the patch length. The token sequence is passed through a transformer encoder \cite{vaswani2017attention} with $L$ layers, $H$ attention heads, and hidden dimension $d$. Full hyperparameters are listed in Appendix~\ref{app:hyperparams}.

\subsection{Data}
We pre-train and evaluate on the ERA5-Land reanalysis \cite{munoz2021era5}, which provides globally consistent daily fields at $0.1^\circ$ spatial resolution by combining numerical weather prediction outputs with assimilated observations (Appendix~\ref{app:data}). We select 21 variables spanning the years 2001--2024, covering near-surface atmospheric and soil components of the water and energy cycles. We treat each $0.1^\circ$ grid cell as a separate entity characterized by 21 co-evolving time series. Together these variables form complementary \textit{views} of the same physical location.
 
To provide spatial and temporal context that is not captured in the meteorological features themselves, we add four positional features to each entity: latitude, longitude, and a sine--cosine encoding of day-of-year:
$\phi_{\text{pos}}(t, \text{lat}, \text{lon}) = \Bigl[\sin\!\left(\tfrac{2\pi t}{365.25}\right),\; \cos\!\left(\tfrac{2\pi t}{365.25}\right),\; \tfrac{\text{lat}}{1800},\; \tfrac{\text{lon}}{3600}\Bigr]$.
These features are not corrupted during pre-training and are not counted toward the 21 features used in the \textit{imposter} task.

We fix the temporal window to 30 days with patch length 2 across all pre-training and downstream experiments. This configuration was selected empirically as a balance between compute cost and downstream task coverage. 

\subsection{Baselines}
We compare \textit{imposter} against five pretext tasks that share its general structure (corruption-based, forecasting, or contrastive objective applied to multivariate temporal inputs):
 
\begin{itemize}
    \item \textbf{Contrastive (NTXent)}: Following \citet{chen2020simple}, we construct positive pairs by sampling two non-overlapping temporal windows from the same entity (same spatial location, different time periods). Other entities in the batch serve as negatives. The InfoNCE objective is applied to the projection of the mean-pooled patch embeddings.
    \item \textbf{Next Token Prediction (NTP)}: Following \cite{radford2019language}, we implement a forecasting objective: given a time series of size $t$ the model is trained to predict all features at the step  $t+1$.
    \item \textbf{Feat-MAE}: A subset of input features (75\%) is selected per sample and masked across all time steps. The model reconstructs the masked feature values via MSE loss. This is the most direct generative counterpart to \textit{imposter} as both operate on full feature dimensions.
 
    \item \textbf{Temp-MAE}: A subset of transformer patches (75\%) is selected per sample and zeroed across all features. The model reconstructs the masked time steps via MSE loss.
 
    \item \textbf{FT-MAE}: 75\% of individual $(t, f)$ cells in the raw input are independently masked. The model reconstructs them via MSE loss.
\end{itemize}
 
All baselines share the same encoder architecture, optimizer settings, and pre-training schedule as \textit{imposter} for fair comparison. 

\subsection{Evaluation}
We evaluate pre-trained representations via two complementary protocols: downstream supervised tasks on frozen encoders, and unsupervised metrics that characterize the geometry of the embedding space.

\textbf{Downstream Tasks I-II: K\"{o}ppen-Geiger Classification.} We probe representations on the K\"{o}ppen-Geiger climate type classification problem, which divides global climates into 5 broad groups and 30 subgroups. For each pre-trained encoder, we extract the last-token embedding and train a linear model on top to predict the K\"{o}ppen class. We report balanced accuracy for both the 5-class (LP-5) and 30-class (LP-30) versions to account for class imbalance.

\textbf{Downstream Tasks III-IV: Hydrology.} Streamflow (SF) is a primary variable in hydrology, measuring the volumetric rate of water carried by rivers. Given a 30-day window of meteorological forcings, we aim to predict SF at the last day. We use catchments from the CAMELS dataset \cite{addor2017camels} for the labels and corresponding ERA5-Land forcings (rather than CAMELS' original inputs) to ensure consistency with the pre-training distribution. We report site-level median $R^2$ across catchments.

As another CAMELS task, we reconstruct basin identity from the weather drivers. The identity of a catchments is represented with a set of 27 numerical features describing topography, climate, and soil which we predict in a multitask fashion. Due to the limited dataset size of this problem (531 basins), we use 5-fold cross-validation for evaluation and all available weather time series for the given location as inputs instead of a single 30 day window. As a final metric we report RMSE on normalized data.

\textbf{Downstream Tasks V-VII: Carbon Fluxes.} Terrestrial carbon fluxes are an important component of the global carbon cycle, and their prediction in a zero-shot transfer learning setting is a challenging ML problem. For our experiments we rely on CarbonFluxBench \cite{rozanov2026carbonbench} and aim to predict three variables representing different types of fluxes: gross primary production (GPP, carbon uptake by plants), ecosystem respiration (RECO, carbon release by plants and microbes), and net ecosystem exchange ($NEE = RECO - GPP$). As a metric, we report site-level median $R^2$ per target.
 
For all regression tasks, we train an ensemble of 5 MLP heads with different initialization seeds and report the mean prediction. This isolates SSL representation quality from MLP initialization variance and yields more stable performance.

\textbf{Representation Quality Metrics.} Downstream evaluation is supervised in its nature and depends on the task structure (different problems leverage different signal in the embedding). To complement it, we report two unsupervised metrics characterizing the embedding itself:
\begin{enumerate}
    \item \textit{RankMe} \cite{garrido2023rankme} estimates the effective rank of the embedding matrix via the smooth entropy of normalized singular values.
    Higher RankMe indicates that the embedding distribution spans more linearly independent directions leading to less redundancy in the representation. 
    \item \textit{$\alpha$ReQ} \cite{agrawal2022alpha} characterizes the spectral decay rate. Following the original work, we fit a power law $\lambda_i \propto i^{-\alpha}$ to the eigenvalues of the embedding covariance matrix and report $\alpha$ as the slope in log--log space. Higher $\alpha$ indicates sharper concentration of variance in main eigenvectors, whereas lower $\alpha$ indicates a flatter, noisier spectrum.
\end{enumerate}
Both metrics are computed on a fixed held-out set of samples unseen during training. 
\section{Results}
\begin{table*}[h]
\small
\centering
\resizebox{\textwidth}{!}{
\begin{tabular}{l cccccccc c cc}
\toprule
 & \multicolumn{2}{c}{\textbf{Linear Probe}} & \multicolumn{2}{c}{\textbf{CAMELS}} & \multicolumn{3}{c}{\textbf{Carbon Flux}} & \textbf{Summary} & & \multicolumn{2}{c}{\textbf{Repr. Quality}} \\
\cmidrule(r){2-3} \cmidrule(r){4-5} \cmidrule(r){6-8} \cmidrule(r){9-9} \cmidrule(l){11-12}
\textbf{Method} & LP-30$\uparrow$ & LP-5$\uparrow$ & SF $R^2\uparrow$ & Basin RMSE$\downarrow$ & GPP $R^2\uparrow$ & RECO $R^2\uparrow$ & NEE $R^2\uparrow$ & \textbf{Mean Rank}$\downarrow$ & & RankMe$\uparrow$ & $\alpha$ReQ$\downarrow$ \\
\midrule
Contrastive       & \textbf{0.859* $\pm$ 0.003} & \textbf{0.927 $\pm$ 0.001} & 0.370 $\pm$ 0.005 & \textbf{0.600 $\pm$ 0.009} & 0.416 $\pm$ 0.031 & 0.300 $\pm$ 0.056 & 0.135 $\pm$ 0.036 & 4.00 & & 75.8$^{\dagger}$ $\pm$ 0.49 & 1.925 $\pm$ 0.058 \\
NTP               & 0.657 $\pm$ 0.003 & 0.840 $\pm$ 0.002 & 0.352 $\pm$ 0.004 & 0.720 $\pm$ 0.014 & 0.489$^{\dagger}$ $\pm$ 0.026 & \textbf{0.447 $\pm$ 0.032} & 0.216$^{\dagger}$ $\pm$ 0.011 & 4.00 & & 51.5 $\pm$ 4.18 & 2.238 $\pm$ 0.078 \\
Feat-MAE          & 0.720 $\pm$ 0.004 & 0.877 $\pm$ 0.001 & 0.378$^{\dagger}$ $\pm$ 0.003 & 0.681$^{\dagger}$ $\pm$ 0.006 & \textbf{0.499 $\pm$ 0.027} & 0.418 $\pm$ 0.029 & 0.211 $\pm$ 0.022 & \textbf{2.79} & & 70.9 $\pm$ 1.60 & 1.913 $\pm$ 0.074 \\
Temp-MAE          & 0.657 $\pm$ 0.042 & 0.850 $\pm$ 0.017 & 0.381 $\pm$ 0.004 & 0.687 $\pm$ 0.009 & 0.472 $\pm$ 0.037 & 0.418 $\pm$ 0.041 & 0.195 $\pm$ 0.028 & 4.21 & & 54.3 $\pm$ 3.35 & 2.179 $\pm$ 0.087 \\
FT-MAE            & 0.633 $\pm$ 0.003 & 0.837 $\pm$ 0.003 & \textbf{0.384 $\pm$ 0.003} & 0.704 $\pm$ 0.014 & 0.472 $\pm$ 0.015 & 0.338 $\pm$ 0.016 & 0.203 $\pm$ 0.005 & 4.86 & & 59.3 $\pm$ 0.29 & 2.044 $\pm$ 0.005 \\
Imp. (ours)       & 0.719 $\pm$ 0.006 & 0.870 $\pm$ 0.002 & 0.366 $\pm$ 0.001 & 0.766 $\pm$ 0.016 & 0.477 $\pm$ 0.015 & 0.404 $\pm$ 0.052 & 0.196 $\pm$ 0.033 & 4.57 & & \textbf{80.4 $\pm$ 0.89} & \textbf{1.442 $\pm$ 0.013} \\
Multi-Imp. (ours) & 0.728$^{\dagger}$ $\pm$ 0.004 & 0.882$^{\dagger}$ $\pm$ 0.001 & 0.347 $\pm$ 0.009 & 0.858 $\pm$ 0.034 & 0.472 $\pm$ 0.010 & 0.422$^{\dagger}$ $\pm$ 0.015 & \textbf{0.223 $\pm$ 0.012} & 3.79$^{\dagger}$ & & 73.4 $\pm$ 0.69 & 1.539$^{\dagger}$ $\pm$ 0.063 \\
\bottomrule
\end{tabular}
}
\caption{Downstream evaluation of individual objectives. Linear probe accuracy reported for 5-class and 30-class Köppen classification. Streamflow (SF), basin attributes, and carbon flux (GPP, RECO, NEE) scores are median $R^2$/RMSE across basins/sites. RankMe and $\alpha$ReQ measure representation quality. All results reported as mean $\pm$ std over 3 seeds. Bold scores represent the best score in the column, whereas $\dagger$ shows the second best in the column. $^*$ indicates the best score achieved across all objectives and experiments throughout this work. }
\label{tab:main_results}
\end{table*}

\subsection{Single Objectives}

We pre-train the two variants of our method---\textit{imposter} and \textit{multi-imposter}---alongside five representative self-supervised objectives on 10M instances sampled from ERA5-Land, stratified by climate type to ensure balanced representation of ecosystems. All methods use the same architecture and pre-training budget, isolating the effect of the pretext objective. Each experiment is repeated with three random seeds, and we report mean and standard deviation (std) throughout.

Table~\ref{tab:main_results} reveals a clear pattern: \textbf{no single self-supervised objective dominates across downstream tasks}. Instead, each objective performs best on tasks aligned with the structural information it is designed to capture, suggesting that different objectives learn complementary rather than universally superior representations.
Performance across downstream tasks closely reflects the inductive bias of each objective. Contrastive learning performs best on tasks requiring discrimination between entities, achieving the highest accuracy on both climate classification tasks (85.9\% on LP-30 and 92.7\% on LP-5) together with the lowest basin attribute RMSE (0.600). This behavior is expected because NTXent explicitly optimizes for instance-level separability. In contrast, \textit{Multi-Imposter} ranks second on both climate classification tasks (72.8\% and 88.2\%), demonstrating that representations learned by modeling physical coherence also preserve substantial information about ecosystem identity, despite not explicitly optimizing for instance discrimination. However, both \textit{imposter} variants perform considerably worse on basin attribute prediction, indicating that physical coherence alone does not capture all of the stable entity-level information emphasized by contrastive learning.

Reconstruction-based objectives excel on streamflow prediction, with FT-MAE ($R^2=0.384$) and Feat-MAE ($R^2=0.378$) achieving the best results. Streamflow depends strongly on accumulated temporal forcing, including precipitation history and snowpack evolution, suggesting that reconstruction objectives are particularly effective when preserving detailed temporal information is most important.

Carbon flux prediction exhibits a different pattern. Here, reconstruction objectives provide only modest advantages, while our methods match or exceed them across all three targets. \textit{Multi-Imposter} achieves the highest NEE $R^2$ among all single objectives (0.223) while remaining competitive on both GPP and RECO. Although Feat-MAE achieves the best GPP performance (0.499) and NTP the best RECO performance (0.447), \textit{Multi-Imposter} ranks among the top three methods across all carbon flux targets. Unlike streamflow prediction, these tasks appear to benefit less from preserving detailed temporal history and more from modeling physically consistent relationships among interacting variables.

Taken together, these results support the central observation of this study. Rather than revealing a universally superior pretext task, they show that downstream performance depends on the alignment between the inductive bias of the self-supervised objective and the structure required by the downstream task. This conclusion is further reflected in the overall mean ranks. Feat-MAE achieves the best average rank (2.79), followed closely by \textit{Multi-Imposter} (3.79), Contrastive (4.00), and NTP (4.00). The narrow spread among these methods indicates that several objectives provide consistently competitive, but differently specialized, representations.

Representation geometry tells a complementary story. \textit{Imposter} produces embeddings with the highest effective rank (RankMe = 80.4) together with the lowest $\alpha$ReQ value (1.442), indicating that its representations span more independent directions than competing methods. However, consistent with prior observations in the SSL literature~\cite{garrido2023rankme}, richer embedding geometry does not directly translate into uniformly better downstream performance, suggesting that representation quality depends not only on embedding diversity but also on alignment with downstream task requirements.

\subsection{Joint Objectives}

The complementary strengths of the single objectives suggest that they capture different aspects of scientific structure. We therefore investigate whether the information learned through physical coherence is complementary to existing pretext tasks using two strategies: (i) concatenating embeddings from independently trained encoders and (ii) jointly optimizing two objectives within a single encoder. Throughout this section we report relative improvements over the corresponding standalone objectives (see Appendix~\ref{app:concat} and~\ref{app:joint} for absolute results).

\subsubsection{Concatenated Embeddings}

Table~\ref{tab:concat_pct} shows that concatenating \textit{imposter} representations consistently improves nearly every baseline (\textbf{$^*$} marks the best score for a given task
across every experiment reported in this work, including standalone
objectives, concatenated representations, and joint training). The largest gains occur on climate classification (up to +18.5\% on LP-30 when combined with FT-MAE) and carbon flux prediction (up to +54.8\% on NEE when combined with Contrastive), indicating that the physical-coherence signal captures information largely absent from reconstruction, forecasting, and contrastive objectives.

Across nearly every baseline, \textit{Multi-Imposter} produces larger improvements than \textit{Imposter}, suggesting that eliminating the single-donor shortcut yields representations that transfer more effectively across objectives. Basin attribute prediction is the only consistent exception, where concatenation increases RMSE by 13.5--27.0\%. This suggests that representations optimized for physical coherence are not always additive when the downstream task primarily depends on preserving stable entity identity.

\begin{table*}[h]
\centering
\tiny
\resizebox{0.75\textwidth}{!}{
\begin{tabular}{l ccccccc}
\toprule
 & \multicolumn{2}{c}{\textbf{Linear Probe}} & \multicolumn{2}{c}{\textbf{CAMELS}} & \multicolumn{3}{c}{\textbf{Carbon Flux}} \\
\cmidrule(r){2-3} \cmidrule(r){4-5} \cmidrule(l){6-8}
\textbf{Method} & LP-30$\uparrow$ & LP-5$\uparrow$ & SF $R^2\uparrow$ & Basin RMSE$\downarrow$ & GPP $R^2\uparrow$ & RECO $R^2\uparrow$ & NEE $R^2\uparrow$ \\
\midrule
Imp. + Contrastive       & \negc{-1.5} & \textbf{\posc{0.1}*} & \textbf{\posc{10.8}*} & \negc{-13.5} & \posc{8.9}  & \posc{8.3}  & \posc{50.4} \\
Multi-Imp. + Contrastive & \negc{-1.7} & \posc{0.1} & \posc{8.4}  & \negc{-18.5} & \posc{5.3}  & \posc{14.7} & \posc{54.8} \\
\addlinespace
Imp. + NTP               & \posc{12.3} & \posc{5.4} & \posc{7.7}  & \negc{-14.0} & \posc{0.6}  & \negc{-9.2} & \posc{4.6}  \\
Multi-Imp. + NTP         & \posc{14.5} & \posc{6.5} & \posc{4.3}  & \negc{-20.3} & \posc{0.2}  & \textbf{\posc{0.7}*}  & \textbf{\posc{10.2}*} \\
\addlinespace
Imp. + Feat-MAE          & \posc{4.7}  & \posc{2.1} & \posc{4.5}  & \negc{-14.5} & \negc{-0.2} & \negc{-3.6} & \posc{2.8}  \\
Multi-Imp. + Feat-MAE    & \posc{5.7}  & \posc{2.9} & \posc{1.9}  & \negc{-27.0} & \posc{0.2}  & \posc{0.2}  & \posc{9.0}  \\
\addlinespace
Imp. + Temp-MAE          & \posc{13.1} & \posc{4.4} & \posc{4.7}  & \negc{-13.7} & \negc{-0.4} & \negc{-1.2} & \posc{2.1}  \\
Multi-Imp. + Temp-MAE    & \posc{15.5} & \posc{5.8} & \posc{1.3}  & \negc{-17.6} & \posc{3.6}  & \posc{2.6}  & \posc{12.8} \\
\addlinespace
Imp. + FT-MAE            & \posc{16.3} & \posc{5.6} & \posc{4.9}  & \negc{-13.5} & \posc{0.6}  & \posc{18.3} & \posc{1.0}  \\
Multi-Imp. + FT-MAE      & \posc{18.5} & \posc{6.9} & \posc{0.5}  & \negc{-23.3} & \posc{2.3}  & \posc{19.5} & \posc{13.8} \\
\bottomrule
\end{tabular}
}
\caption{Percentage improvement of concatenated representations over the base method. Green indicates improvement, red indicates degradation. $^*$ indicates the best score achieved across all objectives and experiments throughout this work. }
\label{tab:concat_pct}
\end{table*}

\subsubsection{Joint Training}

Joint optimization produces smaller average gains than representation concatenation, suggesting that simultaneously optimizing multiple objectives partially sacrifices the complementary information preserved by independently trained encoders.

One result, however, stands out. Although \textit{Multi-Imposter} performs relatively poorly on basin attribute prediction as a standalone objective, combining it with Contrastive reduces basin RMSE by 1.8\% relative to Contrastive alone, yielding the best basin attribute score observed across all experiments, including standalone objectives, concatenated representations, and other joint-training combinations. One possible explanation is that Contrastive emphasizes inter-entity information while \textit{imposter} relies on cross-feature physical consistency, allowing the two objectives to complement one another.

Overall, these experiments reinforce the central hypothesis of this work. Physical coherence represents a complementary source of self-supervision whose inductive bias differs from those of reconstruction, forecasting, and contrastive learning. Rather than replacing existing objectives, \textit{imposter} expands the range of structural information available for representation learning in scientific data.

\begin{table*}[h]
\centering
\tiny
\resizebox{0.75\textwidth}{!}{
\begin{tabular}{l ccccccc}
\toprule
 & \multicolumn{2}{c}{\textbf{Linear Probe}} & \multicolumn{2}{c}{\textbf{CAMELS}} & \multicolumn{3}{c}{\textbf{Carbon Flux}} \\
\cmidrule(r){2-3} \cmidrule(r){4-5} \cmidrule(l){6-8}
\textbf{Method} & LP-30$\uparrow$ & LP-5$\uparrow$ & SF $R^2\uparrow$ & Basin RMSE$\downarrow$ & GPP $R^2\uparrow$ & RECO $R^2\uparrow$ & NEE $R^2\uparrow$ \\
\midrule
Imp. + Contrastive       & \negc{-1.9} & \negc{-0.5} & \posc{1.9}  & \posc{1.5}  & \posc{5.3}  & \posc{3.0}  & \posc{27.4} \\
Multi-Imp. + Contrastive & \negc{-0.9} & \negc{-0.1} & \negc{-2.4} & \textbf{\posc{1.8}*}  & \posc{9.9}  & \negc{-5.0} & \posc{9.6}  \\
\addlinespace
Imp. + NTP               & \posc{22.4} & \posc{8.9}  & \posc{3.7}  & \posc{6.8}  & \posc{3.5}  & \negc{-17.0} & \negc{-0.5} \\
Multi-Imp. + NTP         & \posc{23.1} & \posc{9.4}  & \posc{4.5}  & \posc{4.2}  & \negc{-1.6} & \negc{-10.7} & \negc{-19.9} \\
\addlinespace
Imp. + Feat-MAE          & \posc{9.2}  & \posc{3.4}  & \negc{-1.1} & \negc{-9.8} & \textbf{\posc{4.0}*}  & \negc{-5.3} & \posc{11.8} \\
Multi-Imp. + Feat-MAE    & \posc{9.6}  & \posc{4.2}  & \negc{-4.8} & \negc{-4.6} & \negc{-1.6} & \posc{3.3}  & \negc{-5.7} \\
\addlinespace
Imp. + Temp-MAE & \posc{2.3} & \posc{0.0} & \negc{-3.7} & \posc{2.9} & \negc{-0.2} & \negc{-19.6} & \negc{-9.2} \\
Multi-Imp. + Temp-MAE & \posc{5.8} & \posc{1.2} & \negc{-3.4} & \negc{-1.3} & \negc{-1.3} & \negc{-10.8} & \posc{5.1} \\
\addlinespace
Imp. + FT-MAE & \posc{20.2} & \posc{6.9} & \negc{-3.1} & \posc{3.3} & \posc{4.9} & \posc{12.4} & \negc{-1.5} \\
Multi-Imp. + FT-MAE & \posc{20.4} & \posc{7.2} & \negc{-2.1} & \posc{4.1} & \posc{7.0} & \posc{19.2} & \posc{2.0} \\
\bottomrule
\end{tabular}
}
\caption{Percentage improvement of joint objectives over the base method. Green indicates improvement, red indicates degradation. $^*$ indicates the best score achieved across all objectives and experiments throughout this work.}
\label{tab:joint_pct}
\end{table*}

\subsection{Computational Efficiency}

We measure FLOPs separately for the shared encoder and the objective specific decoder (Appendix~\ref{app:flops}). Considering only the objective-specific decoder, \textit{imposter} and NTP are the least expensive and identical, since both produce ~$1\times F$-dimensional outputs; MAE-family decoders require approximately $\sim$8$\times$ more FLOPs due to per-patch reconstruction, and Contrastive requires approximately $\sim$289$\times$ more due of its projection head and pairwise similarity computations. End-to-end, the shared encoder dominates: \textit{Imposter}, NTP, and all MAE variants fall within 3\% of one another, whereas Contrastive requires approximately $2.05\times$ the total FLOPs.

Combined with the downstream results, these findings position \textit{imposter} as a computationally efficient discriminative objective that provides complementary representations at a cost comparable to reconstruction and forecasting methods, while remaining substantially cheaper than contrastive learning. When combined with Contrastive, the computational cost is naturally dominated by the contrastive objective itself, meaning the value of the combination lies in its empirical performance gains rather than computational efficiency.

\section{Discussion}  \label{sec:disc}
Our experiments show that no self-supervised objective consistently dominates across downstream applications. Instead, performance reflects the inductive bias of each objective: reconstruction-based methods excel when preserving temporal information is most important, contrastive learning is strongest for entity discrimination, and \textit{imposter} targets cross-feature physical coherence. Consequently, the choice of pretext task should be guided by the downstream task rather than the pursuit of a universally superior objective. For foundation models intended to support diverse applications, consistency across tasks may therefore be more valuable than optimizing a single benchmark.

The central contribution of \textit{imposter} is introducing physical coherence as a complementary source of self-supervision. As a standalone objective, \textit{multi-imposter} is the second most consistent method across downstream tasks while requiring compute comparable to NTP and substantially lower than Contrastive. More importantly, combining \textit{imposter} with existing objectives consistently improves performance, indicating that physical coherence provides information largely absent from current self-supervised objectives. Joint training realizes smaller gains than representation concatenation, suggesting that simultaneous optimization partially loses the complementary information preserved by independently learned encoders. Understanding how to better preserve this complementarity is an important direction for future work.

More broadly, our findings suggest that self-supervised objectives should be viewed as mechanisms for exploiting different structural properties of scientific data rather than competitors in the search for a single best objective. Reconstruction emphasizes temporal and spatial regularities, contrastive learning emphasizes instance discrimination, and \textit{imposter} emphasizes cross-feature physical coherence. This perspective suggests that future progress in scientific self-supervision may come from identifying additional sources of scientific structure and developing objectives that exploit them, as well as from more effective ways of combining complementary objectives.

\section{Conclusion}  \label{sec:conclusion}
We introduced \textit{imposter} and its \textit{multi-imposter} variant, discriminative self-supervised pretext tasks that exploit the physical coherence inherent in multivariate scientific time series. By replacing subsets of an entity’s features with physically plausible observations from other entities and training the encoder to identify the incoherent features, the proposed objective encourages representations that capture the cross-feature dependencies governing scientific systems.

Across ERA5-Land pre-training and seven downstream tasks spanning climate classification, hydrology, carbon flux estimation, and representation analysis, \textit{imposter} demonstrates competitive standalone performance while providing information complementary to existing self-supervised objectives. Our systematic comparison further shows that no single pretext task is universally superior. Instead, each objective captures a different aspect of the structure present in scientific data, leading to distinct inductive biases and downstream strengths. These findings suggest that selecting or combining self-supervised objectives should be guided by the intended downstream applications rather than by a single benchmark.

More broadly, we view this work as illustrating a general principle for designing self-supervised learning in scientific domains. Progress need not come solely from adapting objectives originally developed for language and vision. Scientific systems possess distinctive forms of structure—including physical coherence, conservation laws, symmetries, causal relationships, and other domain-specific constraints—that can themselves serve as sources of self-supervision. We hope this work motivates a broader search for such domain-inspired objectives, ultimately enabling more capable, robust, and scientifically grounded foundation models.

\bibliography{ref}

\newpage

\appendix
\begin{table*}[t]
\centering
\caption{Swap schedule and ratio ablation (500K pre-training samples, $T=30$, patch size 2, 3 seeds per configuration). Total rank sums the per-configuration ranks across all seven metrics within each panel; lower is better. Bold marks the best value per column within a panel.}
\label{tab:swaps}
\setlength{\tabcolsep}{2.6pt}
\renewcommand{\arraystretch}{0.9}
\scriptsize
\resizebox{\ifdim\width>\textwidth\textwidth\else\width\fi}{!}{%
\begin{tabular}{@{}llrrrrrrrrrrrrrrrr@{}}
\toprule
& & \multicolumn{8}{c}{\textit{Imposter}} & \multicolumn{8}{c}{\textit{Multi-Imposter}} \\
\cmidrule(lr){3-10} \cmidrule(l){11-18}
\textbf{Sched.} & $r$ & LP-30$\uparrow$ & LP-5$\uparrow$ & SF$\uparrow$ & Basin$\downarrow$ & GPP$\uparrow$ & RECO$\uparrow$ & NEE$\uparrow$ & Rank$\downarrow$ & LP-30$\uparrow$ & LP-5$\uparrow$ & SF$\uparrow$ & Basin$\downarrow$ & GPP$\uparrow$ & RECO$\uparrow$ & NEE$\uparrow$ & Rank$\downarrow$ \\
\midrule
Const. & 0.25 & 0.725 & 0.873 & 0.316 & 0.646 & 0.435 & 0.374 & 0.171 & 73 & 0.724 & 0.875 & 0.357 & 0.645 & 0.449 & 0.419 & 0.212 & 49 \\
 & 0.50 & 0.767 & 0.902 & 0.353 & 0.654 & 0.446 & 0.441 & 0.226 & 36 & 0.761 & 0.891 & 0.343 & 0.642 & 0.448 & 0.433 & 0.217 & 41 \\
 & 0.75 & 0.735 & 0.874 & 0.332 & 0.676 & 0.443 & 0.394 & 0.163 & 66 & 0.778 & \textbf{0.905} & 0.314 & 0.696 & 0.434 & 0.391 & 0.191 & 62 \\
 & 1.00 & 0.651 & 0.837 & \textbf{0.373} & 0.878 & 0.472 & \textbf{0.505} & \textbf{0.248} & 45 & 0.651 & 0.837 & \textbf{0.373} & 0.878 & 0.472 & \textbf{0.505} & \textbf{0.248} & 49 \\
\addlinespace
Linear & 0.25 & 0.568 & 0.816 & 0.312 & 0.712 & 0.433 & 0.468 & 0.188 & 83 & 0.486 & 0.768 & 0.279 & 0.787 & 0.436 & 0.422 & 0.175 & 89 \\
 & 0.50 & 0.656 & 0.850 & 0.327 & 0.717 & 0.428 & 0.391 & 0.191 & 83 & 0.577 & 0.798 & 0.319 & 0.781 & 0.433 & 0.406 & 0.159 & 90 \\
 & 0.75 & 0.689 & 0.867 & 0.363 & 0.695 & \textbf{0.476} & 0.437 & 0.192 & 50 & 0.755 & 0.887 & 0.361 & 0.697 & 0.490 & 0.403 & 0.196 & 46 \\
 & 1.00 & 0.636 & 0.830 & 0.308 & 0.742 & 0.442 & 0.384 & 0.170 & 94 & \textbf{0.779} & 0.903 & 0.341 & 0.733 & 0.467 & 0.407 & 0.215 & 43 \\
\addlinespace
Random & 0.25 & 0.711 & 0.869 & 0.357 & \textbf{0.626} & 0.468 & 0.398 & 0.191 & 47 & 0.708 & 0.868 & 0.360 & 0.636 & 0.496 & 0.400 & 0.220 & 43 \\
 & 0.50 & 0.760 & 0.896 & 0.358 & 0.645 & 0.463 & 0.431 & 0.195 & 36 & 0.730 & 0.879 & 0.345 & 0.686 & 0.437 & 0.368 & 0.160 & 73 \\
 & 0.75 & \textbf{0.776} & \textbf{0.907} & 0.348 & 0.663 & 0.458 & 0.477 & 0.195 & \textbf{33} & 0.763 & 0.892 & 0.356 & 0.635 & 0.480 & 0.344 & 0.211 & 43 \\
 & 1.00 & 0.773 & 0.906 & 0.327 & 0.681 & 0.462 & 0.445 & 0.215 & 38 & 0.774 & 0.899 & 0.335 & \textbf{0.632} & 0.457 & 0.411 & 0.199 & \textbf{40} \\
\addlinespace
R.+L. & 0.25 & 0.528 & 0.798 & 0.293 & 0.758 & 0.438 & 0.476 & 0.187 & 91 & 0.428 & 0.737 & 0.237 & 0.826 & 0.427 & 0.452 & 0.148 & 97 \\
 & 0.50 & 0.539 & 0.804 & 0.289 & 0.746 & 0.426 & 0.476 & 0.191 & 90 & 0.553 & 0.791 & 0.293 & 0.765 & 0.450 & 0.410 & 0.179 & 82 \\
 & 0.75 & 0.700 & 0.869 & 0.360 & 0.690 & 0.465 & 0.415 & 0.195 & 52 & 0.709 & 0.872 & 0.356 & 0.680 & 0.474 & 0.370 & 0.195 & 61 \\
 & 1.00 & 0.756 & 0.892 & 0.365 & 0.645 & 0.470 & 0.388 & 0.205 & 35 & 0.747 & 0.886 & 0.356 & 0.637 & \textbf{0.514} & 0.373 & 0.212 & 44 \\
\bottomrule
\end{tabular}%
}
\end{table*}

\section{Swap Scheduling}
\label{app:swaps}

Table~\ref{tab:swaps} reports the results of our experiments across different \textit{swap schedules} and \textit{swap ratios}. We consider four scheduling strategies:

\begin{itemize}\itemsep0pt \parskip0pt \topsep2pt \partopsep0pt
    \item \textbf{Constant.} The swap ratio is fixed throughout training: the same number of features is replaced for every sample.
    \item \textbf{Linear.} The number of donated features per entity increases linearly with the epoch count.
    \item \textbf{Random.} Given a maximum swap ratio for the run, the individual swap ratio for each instance is sampled uniformly at random.
    \item \textbf{Random+Linear.} Random sampling with a linearly increasing upper bound: $r_t \sim \mathcal{U}(0,\, r_{\max,t})$, where $r_{\max,t} = r_{\max} \cdot \frac{t}{T}$, $t$ is the current epoch, $T$ the total number of epochs, and $r_{\max}$ the maximum swap ratio for the run.
\end{itemize}

\begin{table*}[t]
\centering
\caption{Evaluation of the concatenated representations. Underlined values indicate an improvement over the corresponding base method, and bold values represent the best performance in each column. $^*$ indicates the best score achieved across all objectives and experiments throughout this work.}
\label{tab:concat}
\setlength{\tabcolsep}{4pt}
\renewcommand{\arraystretch}{0.9}
\resizebox{\ifdim\width>\textwidth\textwidth\else\width\fi}{!}{%
\begin{tabular}{@{}lccccccc@{}}
\toprule
 & \multicolumn{2}{c}{\textbf{Linear Probe}} & \multicolumn{2}{c}{\textbf{CAMELS}} & \multicolumn{3}{c}{\textbf{Carbon Flux}} \\
\cmidrule(r){2-3} \cmidrule(r){4-5} \cmidrule(l){6-8}
\textbf{Method} & LP-30$\uparrow$ & LP-5$\uparrow$ & SF $R^2\uparrow$ & Basin RMSE$\downarrow$ & GPP $R^2\uparrow$ & RECO $R^2\uparrow$ & NEE $R^2\uparrow$ \\
\midrule
Imp. + Contrastive        & \textbf{0.846 $\pm$ 0.003} & \underline{\textbf{0.928* $\pm$ 0.001}} & \underline{\textbf{0.410* $\pm$ 0.008}} & \textbf{0.681 $\pm$ 0.010} & \underline{0.453 $\pm$ 0.045} & \underline{0.325 $\pm$ 0.024} & \underline{0.203 $\pm$ 0.035} \\
Multi-Imp. + Contrastive  & 0.844 $\pm$ 0.003 & \underline{\textbf{0.928 $\pm$ 0.001}} & \underline{0.401 $\pm$ 0.007} & 0.711 $\pm$ 0.013 & \underline{0.438 $\pm$ 0.008} & \underline{0.344 $\pm$ 0.026} & \underline{0.209 $\pm$ 0.015} \\
\addlinespace
Imp. + NTP                & \underline{0.738 $\pm$ 0.005} & \underline{0.885 $\pm$ 0.002} & \underline{0.379 $\pm$ 0.006} & 0.821 $\pm$ 0.029 & \underline{0.492 $\pm$ 0.018} & 0.406 $\pm$ 0.035 & \underline{0.226 $\pm$ 0.023} \\
Multi-Imp. + NTP          & \underline{0.752 $\pm$ 0.002} & \underline{0.895 $\pm$ 0.002} & \underline{0.367 $\pm$ 0.006} & 0.866 $\pm$ 0.018 & 0.490 $\pm$ 0.017 & \underline{\textbf{0.450* $\pm$ 0.014}} & \underline{\textbf{0.238* $\pm$ 0.004}} \\
\addlinespace
Imp. + Feat-MAE           & \underline{0.754 $\pm$ 0.002} & \underline{0.895 $\pm$ 0.001} & \underline{0.395 $\pm$ 0.002} & 0.780 $\pm$ 0.037 & 0.498 $\pm$ 0.014 & 0.403 $\pm$ 0.027 & \underline{0.217 $\pm$ 0.008} \\
Multi-Imp. + Feat-MAE     & \underline{0.761 $\pm$ 0.003} & \underline{0.902 $\pm$ 0.002} & \underline{0.385 $\pm$ 0.001} & 0.865 $\pm$ 0.012 & \underline{\textbf{0.500 $\pm$ 0.010}} & \underline{0.419 $\pm$ 0.031} & \underline{0.230 $\pm$ 0.022} \\
\addlinespace
Imp. + Temp-MAE           & \underline{0.743 $\pm$ 0.005} & \underline{0.887 $\pm$ 0.003} & \underline{0.399 $\pm$ 0.005} & 0.781 $\pm$ 0.011 & 0.470 $\pm$ 0.032 & 0.413 $\pm$ 0.009 & \underline{0.199 $\pm$ 0.009} \\
Multi-Imp. + Temp-MAE     & \underline{0.759 $\pm$ 0.002} & \underline{0.899 $\pm$ 0.002} & \underline{0.386 $\pm$ 0.006} & 0.808 $\pm$ 0.025 & \underline{0.489 $\pm$ 0.052} & \underline{0.429 $\pm$ 0.012} & \underline{0.220 $\pm$ 0.010} \\
\addlinespace
Imp. + FT-MAE             & \underline{0.736 $\pm$ 0.005} & \underline{0.884 $\pm$ 0.003} & \underline{0.403 $\pm$ 0.004} & 0.799 $\pm$ 0.008 & \underline{0.475 $\pm$ 0.025} & \underline{0.400 $\pm$ 0.015} & \underline{0.205 $\pm$ 0.006} \\
Multi-Imp. + FT-MAE       & \underline{0.750 $\pm$ 0.002} & \underline{0.895 $\pm$ 0.002} & \underline{0.386 $\pm$ 0.007} & 0.868 $\pm$ 0.017 & \underline{0.483 $\pm$ 0.034} & \underline{0.404 $\pm$ 0.007} & \underline{0.231 $\pm$ 0.013} \\
\bottomrule
\end{tabular}%
}
\end{table*}

\begin{table*}[t]
\centering
\caption{Evaluation of joint objectives. Underlined values indicate an improvement over the corresponding base method, and bold values represent the best performance in each column. $^*$ indicates the best score achieved across all objectives and experiments throughout this work.}
\label{tab:joint}
\setlength{\tabcolsep}{4pt}
\renewcommand{\arraystretch}{0.9}
\resizebox{\ifdim\width>\textwidth\textwidth\else\width\fi}{!}{%
\begin{tabular}{@{}lccccccc@{}}
\toprule
 & \multicolumn{2}{c}{\textbf{Linear Probe}} & \multicolumn{2}{c}{\textbf{CAMELS}} & \multicolumn{3}{c}{\textbf{Carbon Flux}} \\
\cmidrule(r){2-3} \cmidrule(r){4-5} \cmidrule(l){6-8}
\textbf{Method} & LP-30$\uparrow$ & LP-5$\uparrow$ & SF $R^2\uparrow$ & Basin RMSE$\downarrow$ & GPP $R^2\uparrow$ & RECO $R^2\uparrow$ & NEE $R^2\uparrow$ \\
\midrule
Imp. + Contrastive       & \textbf{0.843 $\pm$ 0.003} & 0.922 $\pm$ 0.001 & \underline{\textbf{0.377 $\pm$ 0.007}} & \underline{0.591 $\pm$ 0.004} & \underline{0.438 $\pm$ 0.009} & \underline{0.309 $\pm$ 0.014} & \underline{0.172 $\pm$ 0.026} \\
Multi-Imp. + Contrastive & 0.851 $\pm$ 0.002 & \textbf{0.926 $\pm$ 0.001} & 0.361 $\pm$ 0.010 & \underline{\textbf{0.589* $\pm$ 0.009}} & \underline{0.457 $\pm$ 0.013} & 0.285 $\pm$ 0.032 & \underline{0.148 $\pm$ 0.047} \\
\addlinespace
Imp. + NTP               & \underline{0.804 $\pm$ 0.016} & \underline{0.915 $\pm$ 0.009} & \underline{0.365 $\pm$ 0.005} & \underline{0.671 $\pm$ 0.002} & \underline{0.506 $\pm$ 0.010} & 0.371 $\pm$ 0.004 & 0.215 $\pm$ 0.021 \\
Multi-Imp. + NTP         & \underline{0.809 $\pm$ 0.004} & \underline{0.919 $\pm$ 0.001} & \underline{0.368 $\pm$ 0.009} & \underline{0.690 $\pm$ 0.003} & 0.481 $\pm$ 0.006 & 0.399 $\pm$ 0.006 & 0.173 $\pm$ 0.004 \\
\addlinespace
Imp. + Feat-MAE          & \underline{0.786 $\pm$ 0.008} & \underline{0.907 $\pm$ 0.006} & 0.374 $\pm$ 0.012 & 0.748 $\pm$ 0.012 & \underline{\textbf{0.519* $\pm$ 0.026}} & 0.396 $\pm$ 0.017 & \underline{\textbf{0.236 $\pm$ 0.012}} \\
Multi-Imp. + Feat-MAE    & \underline{0.789 $\pm$ 0.008} & \underline{0.914 $\pm$ 0.003} & 0.360 $\pm$ 0.004 & 0.712 $\pm$ 0.004 & 0.491 $\pm$ 0.013 & \underline{\textbf{0.432 $\pm$ 0.041}} & 0.199 $\pm$ 0.024 \\
\addlinespace
Imp. + Temp-MAE (2/3 seeds) & \underline{0.672 $\pm$ 0.006} & 0.850 $\pm$ 0.003 & 0.367 $\pm$ 0.003 & \underline{0.667 $\pm$ 0.006} & 0.471 $\pm$ 0.036 & 0.336 $\pm$ 0.026 & 0.177 $\pm$ 0.024 \\
Multi-Imp. + Temp-MAE (2/3 seeds) & \underline{0.695 $\pm$ 0.007} & \underline{0.860 $\pm$ 0.003} & 0.368 $\pm$ 0.005 & 0.696 $\pm$ 0.005 & 0.466 $\pm$ 0.022 & 0.373 $\pm$ 0.024 & \underline{0.205 $\pm$ 0.017} \\
\addlinespace
Imp. + FT-MAE (2/3 seeds) & \underline{0.761 $\pm$ 0.009} & \underline{0.895 $\pm$ 0.003} & 0.372 $\pm$ 0.013 & \underline{0.681 $\pm$ 0.010} & \underline{0.495 $\pm$ 0.028} & \underline{0.380 $\pm$ 0.010} & 0.200 $\pm$ 0.034 \\
Multi-Imp. + FT-MAE & \underline{0.762 $\pm$ 0.008} & \underline{0.897 $\pm$ 0.004} & 0.376 $\pm$ 0.004 & \underline{0.675 $\pm$ 0.004} & \underline{0.505 $\pm$ 0.015} & \underline{0.403 $\pm$ 0.007} & \underline{0.207 $\pm$ 0.020} \\
\bottomrule
\end{tabular}%
}
\end{table*}

\section{Model and Training Configuration}
\label{app:hyperparams}

To ensure a fair comparison, all pretext objectives were trained on an identical patch-transformer backbone (only the objective-specific decoder differs between methods). Table~\ref{tab:hyper} lists the architecture and optimization settings shared across all runs.


\begin{table}[H]
\centering
\caption{Shared architecture and optimization hyperparameters.}
\label{tab:hyper}
\scriptsize
\setlength{\tabcolsep}{3pt}
\renewcommand{\arraystretch}{0.9}
\begin{tabular}{@{}ll@{}}
\toprule
\textbf{Parameter} & \textbf{Value} \\
\midrule
Backbone & Patch transformer \\
Embedding dimension $d$ & 128 \\
Transformer blocks & 4 \\
Attention heads & 8 \\
MLP ratio & 4 \\
Projection dimension & 64 \\
Patch size & 2 \\
Dropout & 0.2 \\
\midrule
Optimizer & AdamW \\
Learning rate & $1 \times 10^{-3}$ \\
Weight decay & 0.05 \\
Scheduler & ReduceLROnPlateau \\
Max epochs & 40 \\
Early stopping patience & 5 \\
\midrule
Window size $T$ & 30 \\
Input features $F$ & 21 (+4 positional) \\
Pre-training samples & 10M \\
Seeds per configuration & 3 \\
\bottomrule
\end{tabular}
\end{table}

\section{Dataset}
\label{app:data}

\paragraph{Variables.} We select 21 variables spanning the principal components of the land-surface energy and water cycles, listed in Table~\ref{tab:variables} by physical process group.

\begin{table}[H]
\centering
\caption{ERA5-Land variables used for pre-training, grouped by physical process.}
\label{tab:variables}
\scriptsize
\setlength{\tabcolsep}{3pt}
\renewcommand{\arraystretch}{0.9}
\begin{tabular}{@{}ll@{}}
\toprule
\textbf{Group} & \textbf{Variable} \\
\midrule
\multirow{4}{*}{Temperature} & 2m air temperature \\
 & 2m dewpoint temperature \\
 & Soil temperature (level 1) \\
 & Soil temperature (level 4) \\
\addlinespace
\multirow{3}{*}{Radiation} & Surface net solar radiation \\
 & Surface net thermal radiation \\
 & Surface downward solar radiation \\
\addlinespace
\multirow{2}{*}{Turbulent fluxes} & Surface latent heat flux \\
 & Surface sensible heat flux \\
\addlinespace
\multirow{5}{*}{Water cycle} & Total precipitation \\
 & Total evaporation \\
 & Soil water content (layer 1) \\
 & Soil water content (layer 4) \\
 & Runoff \\
\addlinespace
\multirow{2}{*}{Vegetation} & Leaf area index (high veg.) \\
 & Leaf area index (low veg.) \\
\addlinespace
\multirow{2}{*}{Snow} & Snow cover \\
 & Snow depth water equivalent \\
\addlinespace
\multirow{3}{*}{Atmosphere} & Surface pressure \\
 & 10m wind ($u$ component) \\
 & 10m wind ($v$ component) \\
\bottomrule
\end{tabular}
\end{table}

\paragraph{Sampling.} Pre-training instances are sampled randomly from the global grid with stratification by Köppen--Geiger climate class.

\section{Concatenated embeddings}
\label{app:concat}

Table~\ref{tab:concat} evaluates representations formed by concatenating an \textit{imposter} encoder with each baseline method. Improvements are near-universal across all tasks.

\section{Joint Training}
\label{app:joint}

Table~\ref{tab:joint} repeats this evaluation for encoders trained jointly on both objectives from scratch. Average gains are smaller than under concatenation, suggesting co-optimization discards part of the complementary signal.

\section{Computational Efficiency}
\label{app:flops}

Table~\ref{tab:flops_combined} separates the objective-specific decoder cost from the full forward pass through the shared encoder. 

\begin{table}[H]
\centering
\caption{Task-specific computational overhead (FLOPs) for the head/decoder components vs. Full-Model End-to-End FLOPs (including the shared encoder, $23,784,960$ FLOPs). Overhead metrics are reported relative to imposter.}
\label{tab:flops_combined}
\footnotesize
\setlength{\tabcolsep}{3pt}
\renewcommand{\arraystretch}{0.9}
\begin{tabular}{@{}lrrrr@{}}
\toprule
 & \multicolumn{2}{c}{\textbf{Decoder Only}} & \multicolumn{2}{c}{\textbf{Encoder+Decoder}} \\
\cmidrule(r){2-3} \cmidrule(l){4-5}
\textbf{Method} & \textbf{FLOPs} & \textbf{Overhead} & \textbf{FLOPs} & \textbf{Overhead} \\
\midrule
Imp.        &     87,296 &   1.00$\times$ & 23,872,256 & 1.00$\times$ \\
NTP         &     87,296 &   1.00$\times$ & 23,872,256 & 1.00$\times$ \\
Feat-MAE    &    701,952 &   8.04$\times$ & 24,486,912 & 1.03$\times$ \\
Temp-MAE    &    701,952 &   8.04$\times$ & 24,486,912 & 1.03$\times$ \\
FT-MAE      &    701,952 &   8.04$\times$ & 24,486,912 & 1.03$\times$ \\
Contrastive & 25,188,992 & 288.55$\times$ & 48,973,952 & 2.05$\times$ \\
\bottomrule
\end{tabular}
\end{table}
\end{document}